\documentclass[conference]{IEEEtran}
\IEEEoverridecommandlockouts

\usepackage{amsmath,amssymb,amsfonts}
\usepackage{algorithmicx}
\usepackage{graphicx}
\usepackage{textcomp}
\usepackage{xcolor}
\usepackage[pagebackref=true,breaklinks=true,letterpaper=true,colorlinks,bookmarks=false]{hyperref}
\usepackage[nocompress]{cite}
\usepackage{multicol}
\usepackage{url}
\usepackage{ragged2e}
\usepackage{epsfig}
\usepackage{mathptmx}      
\usepackage{subfigure}
\usepackage{algorithm}
\usepackage{algpseudocode}
\usepackage{graphics}
\usepackage{mathrsfs}
\usepackage{threeparttable}
\usepackage{color}
\usepackage[normalem]{ulem}
\usepackage{multirow}
\usepackage{float}
\usepackage{amsfonts}
\usepackage{bm}
\usepackage{array}
\usepackage{colortbl}
\usepackage{pifont}
\usepackage{diagbox}
\usepackage{rotating}
\usepackage{booktabs}
\usepackage{overpic}
\usepackage{contour}
\usepackage{enumitem}
\usepackage{stfloats}
\usepackage{threeparttable}
\usepackage{makecell}

\ifCLASSINFOpdf
\else
   \usepackage[dvips]{graphicx}
\fi
\usepackage{url}

\newcommand{\figref}[1]{Fig.~\ref{#1}}
\newcommand{\tabref}[1]{Table~\ref{#1}}

\def\eg{\textit{e.g.}}

\def\BibTeX{{\rm B\kern-.05em{\sc i\kern-.025em b}\kern-.08em T\kern-.1667em\lower.7ex\hbox{E}\kern-.125emX}}
\begin{document}

\title{Promoting Segment Anything Model towards Highly Accurate Dichotomous Image Segmentation}

\author{
	\IEEEauthorblockN{
		Xianjie Liu$^a$, 
		Keren Fu$^a$\IEEEauthorrefmark{1}, 
		Yao Jiang$^a$, 
		and Qijun Zhao$^a$} 
	\IEEEauthorblockA{$^a$College of Computer Science, Sichuan University, Sichuan, China}
	\IEEEauthorblockA{\IEEEauthorrefmark{1}Corresponding emails: fkrsuper@scu.edu.cn}
} 

\maketitle

\begin{abstract}
The Segment Anything Model (SAM) represents a significant breakthrough into foundation models for computer vision, providing a large-scale image segmentation model. However, despite SAM's zero-shot performance, its segmentation masks lack fine-grained details, particularly in accurately delineating object boundaries. Therefore, it is both interesting and valuable to explore whether SAM can be improved towards highly accurate object segmentation, which is known as the dichotomous image segmentation (DIS) task. To address this issue, we propose DIS-SAM, which advances SAM towards DIS with extremely accurate details. DIS-SAM is a framework specifically tailored for highly accurate segmentation, maintaining SAM's promptable design. DIS-SAM employs a two-stage approach, integrating SAM with a modified advanced network that was previously designed to handle the prompt-free DIS task. To better train DIS-SAM, we employ a ground truth enrichment strategy by modifying original mask annotations. Despite its simplicity, DIS-SAM significantly advances the SAM, HQ-SAM, and Pi-SAM by $\sim$8.5\%, $\sim$6.9\%, and $\sim$3.7\% maximum F-measure. Our code at \href{https://github.com/Tennine2077/DIS-SAM}{DIS-SAM}.

\end{abstract}


\section{Introduction}
\label{sec:intro}

The Segment Anything Model (SAM) \cite{kirillov2023segment} is a significant breakthrough in computer vision foundation models, aiming to solve the long-standing image segmentation problem with versatility and scalability. SAM is designed as a large-scale model that can take images and various promptable segmentation queries (\eg, points, bounding boxes) as inputs, enabling users to guide the segmentation process interactively. One of SAM's key strengths is its impressive zero-shot performance without requiring task-specific training, SAM can generalize across a wide range of segmentation tasks, showing robust results on unseen data.

Since its debut in 2023, SAM has gained significant attention, amassing over 7.4k Google Scholar citations and 48.2k GitHub stars, highlighting its impact. SAM can be applied in many fields, such as camouflaged object detection \cite{liang2024finet}, medical image segmentation \cite{zheng2024polyp}, and few-shot semantic segmentation \cite{jiang2024prototypical}. Additionally, SAM2\cite{ravi2024sam2} extends SAM's functionality by adding video segmentation capabilities.

\begin{figure}
    \centering
    \includegraphics[width=0.48\textwidth]{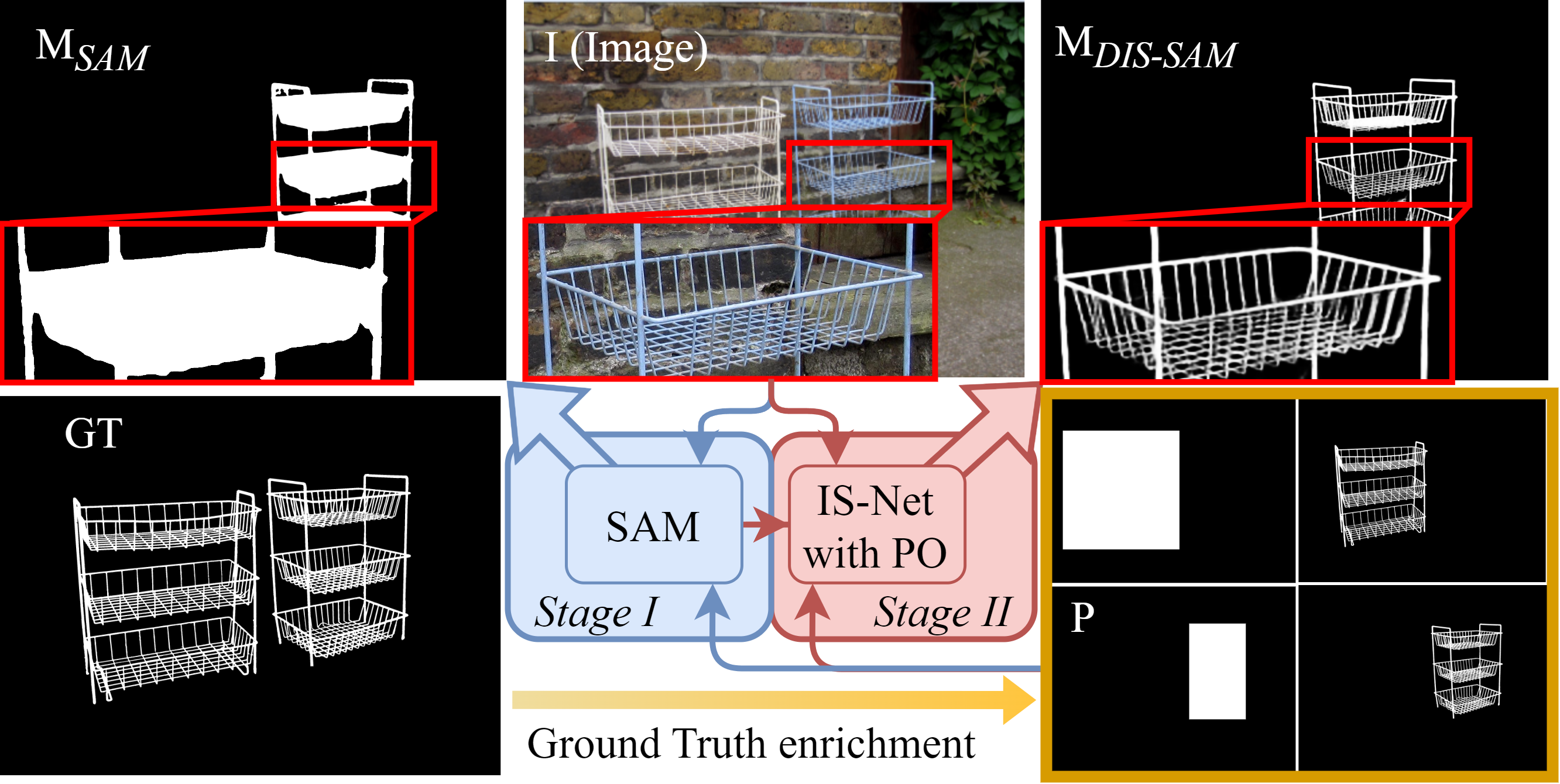}
    \caption{Overall pipeline of the proposed DIS-SAM.}
    \vspace{-0.35cm}
    \label{fig:DIS-SAM}
\end{figure}

Despite these achievements, SAM faces limitations when it comes to fine-grained segmentation. While it excels in identifying and segmenting general object regions, the segmentation masks it produces are often coarse, with blurry or inaccurate object boundaries. This becomes particularly problematic in applications where high precision is critical, such as medical diagnostics, detailed image editing, or tasks involving thin or intricate objects \cite{zhang2023comprehensive}. The masks generated by SAM tend to lack necessary details for such scenarios (e.g., \figref{fig:DIS-SAM}), as they fail to capture subtle or complex boundary features.

Efforts have been made to mitigate this issue, most notably with the development of HQ-SAM \cite{ke2024segment}, which utilizes prompt learning, and Pi-SAM \cite{mengzhen2024segment}, which utilizes an additional embedder and decoder.
The above advances have made progress over the original SAM, they still fall short in delivering the level of precision required for highly accurate segmentation tasks, particularly when dealing with fine structures.

This inadequacy is especially evident in the context of dichotomous image segmentation (DIS) \cite{qin2022highly}, a task that demands fine object boundaries and pixel-perfect accuracy, essential for applications like art design, product editing, and scientific imaging. However, previous DIS methods like \cite{qin2022highly,pei2023unite,zheng2024birefnet,zhou2023dichotomous,yu2024multi} have been prompt-free, restricting flexibility to automatic, non-interactive segmentation of primary objects and limiting adaptability to tasks requiring user input or guidance.

Given these challenges, it is natural to ask: \textit{Can SAM be promoted to achieve highly accurate dichotomous image segmentation (DIS) while preserving its interactive, promptable design?} To this end, we introduce \textbf{DIS-SAM} 
(as shown in \figref{fig:DIS-SAM})
a framework designed to push SAM towards highly accurate, fine-grained object segmentation while maintaining the flexibility of user prompts. We adopt a two-stage approach, directly connecting the output of SAM to the input of IS-Net. The latter is an advanced network that was previously designed in \cite{qin2022highly} to tackle the prompt-free DIS task.
By refining the rough segmentation masks generated by SAM using IS-Net, our method effectively addresses issues such as coarse boundaries and inaccurate segmentation from SAM alone. Leveraging its architectural features and pre-trained ground truth (GT)  encoder for feature supervision, IS-Net corrects rough boundaries to produce fine-grained object edges, improving segmentation accuracy. This approach requires no significant modifications to SAM or IS-Net\cite{qin2022highly}, simplifying workflows and reducing development costs. 
In summary, our contributions are as follows:
\begin{itemize}
    \item We introduce a two-stage DIS-SAM framework that integrates SAM’s promptable segmentation with IS-Net, a model dedicated to DIS, significantly improving segmentation accuracy and boundary precision.
    \item A novel data enrichment strategy is proposed to augment the training dataset, effectively enhancing the model’s performance on complex segmentation tasks that involve thin objects or detailed structures.
    \item 
    DIS-SAM significantly advances the SAM, HQ-SAM, and Pi-SAM by $\sim$8.5\%, $\sim$6.9\%, and $\sim$3.7\% maximum F-measure. It also demonstrates robust performance and strong generalization capabilities in zero-shot scenarios across multiple datasets, making it highly effective for accurate segmentation tasks.
\end{itemize}

\section{Related Work}
\subsection{Segment Anything Models (SAM)}

SAM\cite{kirillov2023segment} is a foundation model for image segmentation, with many downstream tasks relying on its performance, driving efforts to improve its accuracy.
HQ-SAM\cite{ke2024segment} improves segmentation precision by introducing a learnable high-quality output token into SAM's mask decoder, which significantly refines boundary predictions. To bolster generalization, HQ-SAM incorporates HQSeg-44K, a dataset featuring 44K refined masks collected from various sources.
Pi-SAM\cite{mengzhen2024segment} enhances output mask accuracy by adopting a high-resolution mask decoder and provides an optional precision interactor. The high-resolution mask decoder ensures finer segmentation, while the precision interactor allows users to refine predictions interactively through clicks, addressing errors effectively.
However, despite the significant progress made by HQ-SAM and Pi-SAM compared to the original SAM, they still exhibit limitations when addressing DIS tasks.

\subsection{Dichotomous Image Segmentation (DIS)}
DIS is a high-precision segmentation task requiring meticulous object boundary delineation and detailed accuracy. 
After a specific high-resolution DIS dataset was proposed, IS-Net \cite{qin2022highly} became the first work targeting the DIS task. It utilized $U^2$-Net \cite{qin2020u2} and a GT encoder for intermediate supervision to alleviate the loss of fine areas and achieved good results. UDUN \cite{pei2023unite} proposed a unite-divide-unite approach. It conducted segmentation by decoupling the trunk and edges and achieved excellent segmentation performance on the boundaries of objects. BiRefNet \cite{zheng2024birefnet} proposed a bilateral reference strategy, taking the original image patches and image edges as internal and external references. It utilized intact high-resolution images as supplementary information and improved the segmentation precision. 
However, existing methods tailored to DIS share a notable limitation: they are prompt-free. This absence of user interaction or input adaptability constrains their applicability to tasks that require user guidance, such as image editing. Consequently, these approaches are restricted to automatically segmenting primary objects without interactivity, limiting their versatility across diverse scenarios.

\section{Proposed Method}

The proposed DIS-SAM is a two-stage framework designed to improve the Segment Anything Model (SAM) for highly accurate dichotomous image segmentation (DIS), as shown in \figref{fig:DIS-SAM}. DIS-SAM enhances SAM by introducing a fine-grained segmentation stage using IS-Net while retaining SAM’s promptable features. Below, we detail each component of the method, starting from the overall architecture to specific training strategies, including parameter orthogonalization (PO), loss function design, and GT enrichment.

\subsection{Model Pipeline}

The DIS-SAM pipeline in \figref{fig:DIS-SAM} is composed of two main stages. Let $\mathbf{I} \in \mathbb{R}^{H \times W \times 3}$ represent the input image of height $H$ and width $W$, and $\mathbf{P}$ denote the user-provided prompt (bounding box). In the first stage, SAM takes both $\mathbf{I}$ and $\mathbf{P}$ as inputs and generates a coarse segmentation mask $\mathbf{M}_{SAM} \in \mathbb{R}^{H \times W}$:
$\mathbf{M}_{SAM} = \text{SAM}(\mathbf{I}, \mathbf{P})$.
SAM consists of a Vision Transformer (ViT) image encoder $\mathcal{E}_{img}$, a prompt encoder $\mathcal{E}_{prompt}$, and a mask decoder $\mathcal{D}_{mask}$. The image encoder extracts feature maps from the input image $\mathbf{I}$, while the prompt encoder encodes $\mathbf{P}$ to guide the segmentation. The outputs from both encoders are then passed to the mask decoder, generating the coarse mask $\mathbf{M}_{SAM}$. As SAM's architecture is very famous, it details ($\mathcal{E}_{img}$, $\mathcal{E}_{prompt}$ and $\mathcal{D}_{mask}$) are omitted in this paper. 

In the second stage, we concatenate the original image $\mathbf{I}$, the prompt $\mathbf{P_{box}}$, and the coarse mask $\mathbf{M}_{SAM}$ to form a five-channel input tensor $\mathbf{X}_{input}$, namely
$\mathbf{X}_{input} = \text{Concat}(\mathbf{I}, \mathbf{M}_{SAM}, \mathbf{P}_{box})$,
where $\mathbf{P}_{box} \in \mathbb{R}^{H \times W}$ is a binary mask box derived from the user prompt, assigning a value of 1 to pixels inside the prompt region and 0 otherwise. The tensor $\mathbf{X}_{input} \in \mathbb{R}^{H \times W \times 5}$ is then passed to IS-Net, to refine the segmentation mask as
$\mathbf{M}_{DIS\text{-}SAM} = \text{IS-Net}(\mathbf{X}_{input})$.
IS-Net utilizes previous $U^2$-Net \cite{qin2020u2} as the main architecture, and adopts a pre-trained ground truth (GT) encoder to provide intermediate feature supervision during the training phase, enforcing the segmentation model's intermediate features to align with those from the GT encoder. More details of IS-Net are also omitted here and can be referred to \cite{qin2022highly}. The output $\mathbf{M}_{DIS\text{-}SAM}$ is a highly accurate segmentation mask that refines the coarse boundaries of $\mathbf{M}_{SAM}$.

\begin{table*}
\centering
\caption{Performance comparisons of DIS-SAM with IS-Net, UDUN, BiRefNet, SAM, HQ-SAM and Pi-SAM. Since the source code of Pi-SAM is not publicly available, we directly cite its metrics from \cite{mengzhen2024segment}. The symbols ↑/↓ indicate that higher/lower scores are better.}
\label{tab:table1}
\small
\renewcommand{\arraystretch}{1}
\renewcommand{\tabcolsep}{0.5mm}
\begin{tabular}{c|cccccc|cccccc|cccccc}
    \toprule
    Datasets & \multicolumn{6}{c|}{DIS-VD} & \multicolumn{6}{c|}{DIS-TE1} & \multicolumn{6}{c}{DIS-TE2} \\
    \hline
    Metric & $F^{max}_\beta\uparrow$&$F^w_\beta\uparrow$&$~M~\downarrow$&$S_{\alpha}\uparrow$&$E_{\phi}^{m}\uparrow$&$HCE_\gamma\downarrow$ & $F^{max}_\beta\uparrow$&$F^w_\beta\uparrow$&$~M~\downarrow$&$S_{\alpha}\uparrow$&$E_{\phi}^{m}\uparrow$&$HCE_\gamma\downarrow$ & $F^{max}_\beta\uparrow$&$F^w_\beta\uparrow$&$~M~\downarrow$&$S_{\alpha}\uparrow$&$E_{\phi}^{m}\uparrow$&$HCE_\gamma\downarrow$ \\
    \hline
    IS-Net & 0.791 & 0.717 & 0.074 & 0.813 & 0.856 & 1116 & 0.740 & 0.662 & 0.074 & 0.787 & 0.820 & 149 & 0.799 & 0.728 & 0.070 & 0.823 & 0.858 & 340 \\
    UDUN & 0.823 & 0.763 & 0.059 & 0.838 & 0.892 & 1097 & 0.784 & 0.720 & 0.059 & 0.817 & 0.860 & 140 & 0.829 & 0.768 & 0.058 & 0.843 & 0.886 & 325\\
    BiRefNet & 0.891 & 0.854 & 0.038 & 0.898 & 0.931 & 989 & 0.860 & 0.819 & 0.037 & 0.885 & 0.911 & \textbf{106} & 0.894 & 0.857 & 0.036 & 0.900 & 0.930 & \textbf{266}\\
    \hline
    SAM & 0.835 & 0.782 & 0.069 & 0.808 & 0.889 & 1516 & 0.838 & 0.807 & 0.047 & 0.843 & 0.805 & 266 & 0.803 & 0.758 & 0.081 & 0.792 & 0.863 & 582 \\
    HQ-SAM & 0.851 & 0.829 & 0.045 & 0.848 & 0.919 & 1386 & 0.903 & 0.888 & 0.019 & 0.907 & 0.959 & 196 & 0.895 & 0.874 & 0.029 & 0.883 & 0.950 & 466\\
    Pi-SAM & 0.883 & 0.866 & 0.035 & 0.889 & 0.945 & 1322 & 0.890 & 0.869 & 0.027 & 0.894 & 0.947 & 176 & 0.903 & 0.887 & 0.027 & 0.907 & 0.953 & 383 \\
    \textbf{DIS-SAM} & \textbf{0.920} & \textbf{0.877} & \textbf{0.031} & 0.909 & \textbf{0.948} & \textbf{987} & 0.929 & \textbf{0.897} & \textbf{0.019} & 0.929 & \textbf{0.960} & 115 & \textbf{0.924} & \textbf{0.889 }& \textbf{0.025} & \textbf{0.921} & \textbf{0.955} & 287 \\
    \textbf{DIS-SAM$^{*}$} & 0.917 & 0.854 & 0.037 & \textbf{0.910} & 0.931 & 1045 & \textbf{0.939} & 0.881 & 0.024 & \textbf{0.931} & 0.946 & 126 & \textbf{0.923} & 0.870 & 0.032 & \textbf{0.921} & 0.938 & 306 \\
    \hline
    Datasets & \multicolumn{6}{c|}{DIS-TE3} & \multicolumn{6}{c|}{DIS-TE4} & \multicolumn{6}{c}{DIS-TE (ALL)} \\
    \hline
    Metric & $F^{max}_\beta\uparrow$&$F^w_\beta\uparrow$&$~M~\downarrow$&$S_{\alpha}\uparrow$&$E_{\phi}^{m}\uparrow$&$HCE_\gamma\downarrow$ & $F^{max}_\beta\uparrow$&$F^w_\beta\uparrow$&$~M~\downarrow$&$S_{\alpha}\uparrow$&$E_{\phi}^{m}\uparrow$&$HCE_\gamma\downarrow$ & $F^{max}_\beta\uparrow$&$F^w_\beta\uparrow$&$~M~\downarrow$&$S_{\alpha}\uparrow$&$E_{\phi}^{m}\uparrow$&$HCE_\gamma\downarrow$ \\
    \hline
    IS-Net &0.830 & 0.758 & 0.064 & 0.836 & 0.883 & 687 & 0.827 & 0.753 & 0.072 & 0.830 & 0.870 & 2888 & 0.799 & 0.725 & 0.070 & 0.819 & 0.858 & 1016 \\
    UDUN & 0.865 & 0.809 & 0.050 & 0.865 & 0.917 & 658 & 0.846 & 0.792 & 0.059 & 0.849 & 0.901 & 2785 & 0.831 & 0.772 & 0.057 & 0.844 & 0.891 & 977\\
    BiRefNet & \textbf{0.925} & \textbf{0.893} & \textbf{0.028} & 0.919 & \textbf{0.955} & \textbf{569} & \textbf{0.904} & \textbf{0.864} & \textbf{0.039} & 0.900 & \textbf{0.939} & 2723 & 0.896 & 0.858 & 0.035 & 0.901 & 0.934 & 916\\
    \hline
    SAM & 0.773 & 0.724 & 0.094 & 0.761 & 0.848 & 1050 & 0.677 & 0.634 & 0.162 & 0.697 & 0.762 & 3505 & 0.773 & 0.731 & 0.096 & 0.773 & 0.845 & 1351 \\
    HQ-SAM & 0.860 & 0.853 & 0.045 & 0.851 & 0.926 & 927 & 0.776 & 0.748 & 0.088 & 0.799 & 0.863 & 3386 & 0.859 & 0.835 & 0.045 & 0.860 & 0.924 & 1244\\
    Pi-SAM & 0.899 & 0.882 & 0.030 & 0.901 & 0.953 & 779 & 0.869 & 0.855 & 0.046 & 0.871 & \textbf{0.939} & 3299 & 0.890 & 0.873 & 0.033 & 0.893 & 0.948 & 1191\\
    \textbf{DIS-SAM} & 0.918 & 0.877 & 0.030 & 0.908 & 0.948 & 598 & 0.899 & 0.849 & 0.043 & 0.888 & 0.932 & \textbf{2609} & \textbf{0.917} & \textbf{0.878} & \textbf{0.029} & 0.911 & \textbf{0.949} & \textbf{902} \\
    \textbf{DIS-SAM$^{*}$} & 0.913 & 0.860 & 0.035 & \textbf{0.935} & 0.904 & 644 & 0.890 & 0.818 & 0.054 & \textbf{0.904} & 0.931 & 2788 & 0.916 & 0.857 & 0.036 & \textbf{0.912} & 0.930 & 966 \\
    \bottomrule
\end{tabular}
\vspace{-0.32cm}
\end{table*}

\subsection{Parameter Orthogonalization}
To further enhance the generalization ability of IS-Net, especially when trained on a smaller dataset where overfitting is more pronounced, we introduce a parameter orthogonalization (PO) term, namely ORTHO loss \cite{brock2017neural}, which enforces the orthogonality of convolutional filters. Let \( \mathbf{W}_l \) denote the weight matrix of the \( l \)-th convolutional layer, flattened for each filter. The ORTHO loss for layer \( l \) is defined as:
$\left\|\mathbf{W}_l \mathbf{W}_l^\top - \mathbf{E}\right\|_F$, 
where $\|\cdot\|_F$ denotes the Frobenius norm and $\mathbf{E}$ is the identity matrix. This loss is summed over all convolutional layers in IS-Net, where \( L \) represents the total number of convolutional layers:
$L_{ortho} = \sum_{l=1}^{L} \left\|\mathbf{W}_l \mathbf{W}_l^\top - \mathbf{E}\right\|_F$.
By enforcing orthogonality, we reduce redundancy among the filters and improve the model's robustness across different datasets. We find that this technique is able to improve S-measure ($S_{\alpha}$) \cite{fan2017structure} by $\sim$6.6\% while maintaining other evaluation score metrics.

\subsection{Overall Loss Design}
To improve segmentation accuracy, we design a composite loss function that combines binary cross-entropy (BCE) and intersection-over-union (IoU) losses. Let \( \mathbf{M}_{DIS\text{-}SAM} \) and \( \mathbf{M}_{GT} \) represent the predicted mask and the ground truth (GT) mask, respectively. Here, \( p_i \) denotes the predicted probability for the \( i \)-th pixel, \( y_i \) represents the ground truth label for the \( i \)-th pixel, and \( M \) is the total number of pixels in the mask. The BCE loss can be simplified as follows:
\begin{equation}
L_{BCE} = -\frac{1}{M} \sum_{i=1}^{M} \left[ y_i \log(p_i) + (1 - y_i) \log(1 - p_i) \right].
\end{equation}\noindent
The IoU loss, which focuses on the overlap between the predicted and ground truth (GT) masks, is defined as:
\begin{equation}
L_{IoU} = 1 - \frac{\sum_{i=1}^{M} (p_i \cdot y_i)}{\sum_{i=1}^{M} (p_i + y_i) - \sum_{i=1}^{M} (p_i \cdot y_i)}.
\end{equation}\noindent
Therefore, the overall loss function $\mathcal{L}_{\text{total}}$ when training the main architecture is a weighted sum of four terms:
\begin{equation}
\mathcal{L}_{\text{total}} = \lambda_1 L_{\text{BCE}} + \lambda_2 L_{\text{IoU}} + \lambda_3 L_{\text{MSE}} + \lambda_4 L_{\text{ortho}},
\end{equation}\noindent
where $L_{\text{MSE}}$ is the loss used to align intermediate features with the high-dimensional features from GT encoder \cite{qin2020u2}, and is implemented by mean squared error loss. $L_{\text{BCE}}$ and $L_{\text{IoU}}$ are the aforementioned BCE and IoU losses, respectively.
$\lambda_1$, $\lambda_2$ and $\lambda_3$ are empirically set weights at 20, 0.5, and 1 to keep the losses at the same magnitude level. $\lambda_4$ is set to $10^{-6}$.

\subsection{Ground Truth (GT) Enrichment}
\begin{figure}
  \centering
  \includegraphics[width=0.48\textwidth]{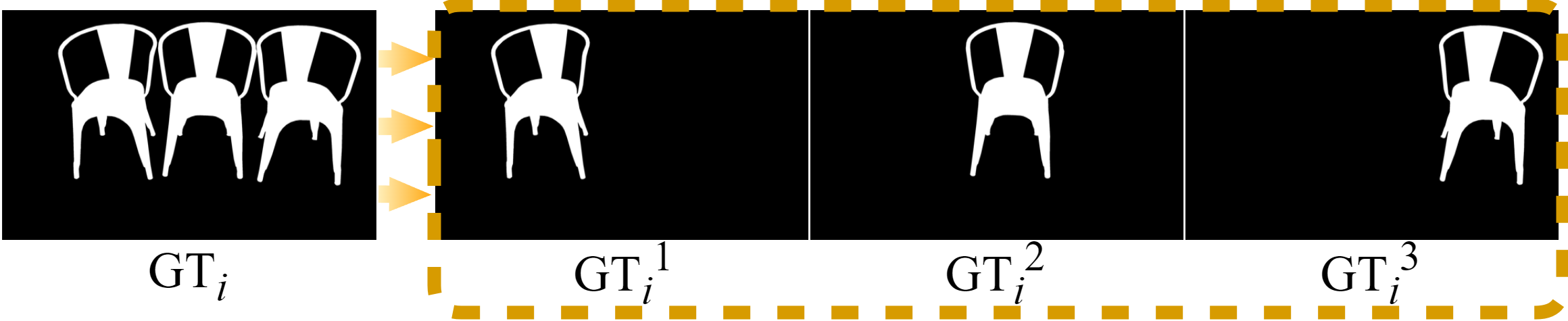}
  \caption{
  An example of segmenting out connected components, where the GT image is decomposed into three parts, corresponding to three masks. For the sake of space, the original color image is omitted.
 }
 \vspace{-0.32cm}
  \label{fig:Enrichment}
\end{figure}

In order to adapt to a promptable mode, we further employ a data enrichment strategy as shown in \figref{fig:Enrichment}, by modifying the GT annotations. Let $\mathbf{GT}_i \in \mathbb{R}^{H \times W}$ represent the GT mask for image $i$. If $\mathbf{GT}_i$ contains multiple disjoint objects, we split it into $N$ non-overlapping regions $\{\mathbf{GT}_i^1, \mathbf{GT}_i^2, \dots, \mathbf{GT}_i^N\}$. Each $\mathbf{GT}_i^n$ corresponds to an individual object in the image, and thus each object forms a new image-GT pair: $\{\mathbf{I}_i, \mathbf{GT}_i^n\}$, for all $n = 1, 2, \dots, N$. This data enrichment method increases the number of training samples, allowing DIS-SAM to generalize to different objects and scenarios.

\begin{figure*}
  \centering
  \includegraphics[width=0.96\textwidth]{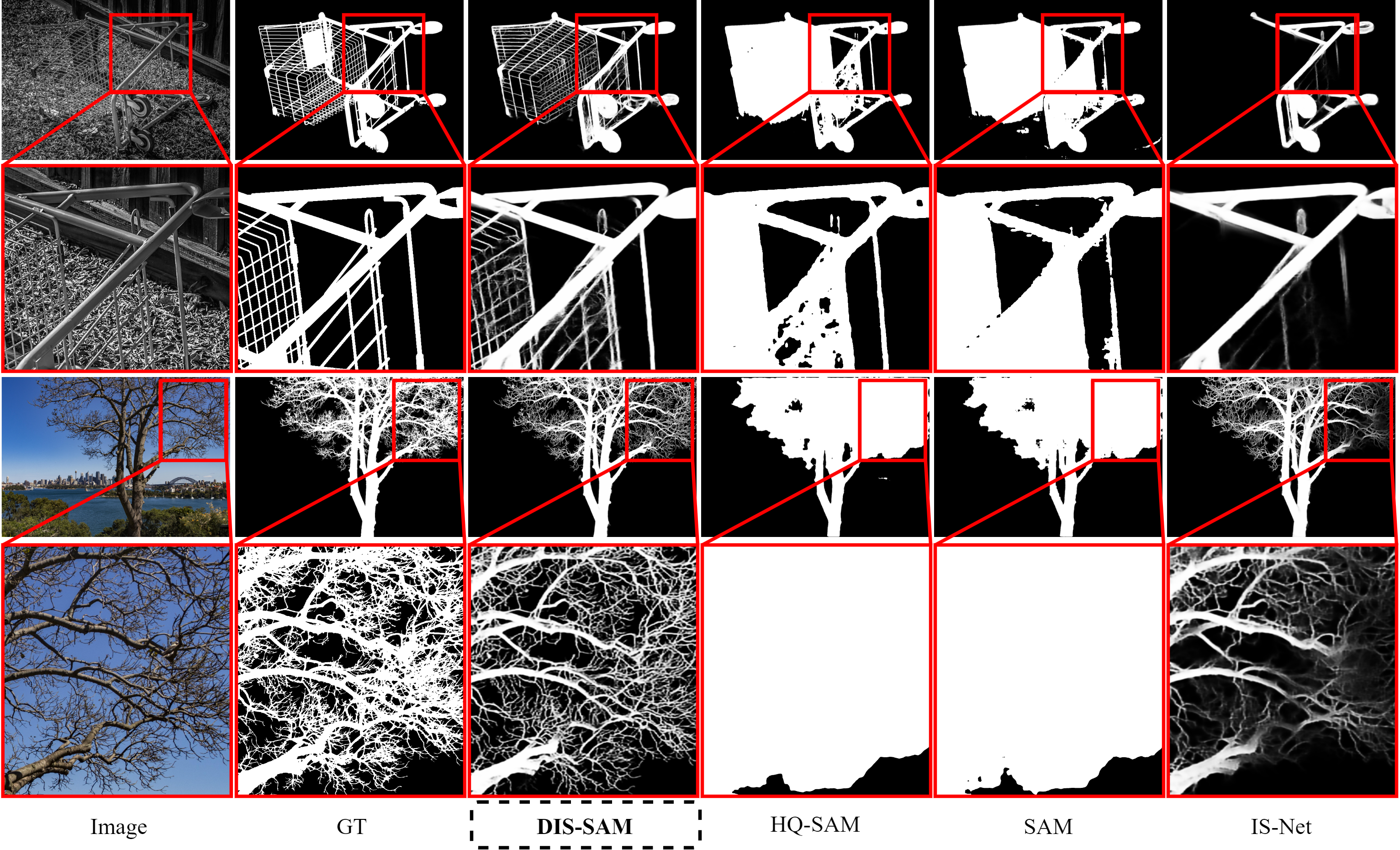}
  \caption{Visual results of DIS-SAM, HQ-SAM \cite{ke2024segment}, SAM \cite{kirillov2023segment}, and IS-Net \cite{qin2022highly}.}
  \vspace{-0.35cm}
  \label{fig:compare}
\end{figure*}

\section{Experiments and Results}
\subsection{Setups and Implementation Details}
DIS-5K dataset is used for training and evaluation, which consists of 3,000 samples for training, 470 for validation (DIS-VD), and 2,000 for testing (partitioned into four subsets DIS-TE1$\sim$DIS-TE4). After data enrichment, 3,880 samples are available for training.
Additionally, to validate the model's performance and generalization, we used the HQ-SAM's training dataset HQSeg-44k \cite{ke2024segment} to train DIS-SAM, resulting in another variant DIS-SAM$^{*}$. HQSeg-44k combines six datasets with fine-grained annotations, including DIS-5K \cite{qin2022highly} (training set), ThinObject-5K\cite{liew2021deep} (training set), FSS-1000 \cite{li2020fss}, ECSSD \cite{shi2015hierarchical}, MSRA-10K \cite{cheng2014global} and DUT-OMRON \cite{yang2013saliency}. Each provides $\sim$7.4K mask annotations for training. Following \cite{ke2024segment}, zero-shot evaluation is conducted using the test sets of COIFT\cite{liew2021deep}, HRSOD\cite{zeng2019towards} and ThinObject-5K\cite{liew2021deep}.

Following \cite{qin2022highly}, we adopt five metrics, including maximum F-measure ($F^{max}_\beta$) \cite{achanta2009frequency}, weighted F-measure ($F^w_\beta$), mean absolute error ($M$), S-measure ($S_{\alpha}$) \cite{fan2017structure}, average enhanced alignment measure ($E_{\phi}^{m}$) \cite{fan2018enhanced}, and human correction efforts ($HCE_\gamma$) \cite{qin2022highly}, to evaluate from various perspectives. The symbols $\uparrow$/$\downarrow$ in \tabref{tab:table1} indicate higher/lower scores are better.

The DIS-SAM model was trained by freezing SAM's pre-trained weights and fine-tuning only the subsequent IS-Net. During training and testing, all images were resized to $1024^2$. Prompt boxes were generated from ground truth and used to produce coarse segmentation masks with SAM. Data augmentation was limited to horizontal flipping. The training used the Adam optimizer with an initial learning rate of $10^{-5}$, a batch size of 6, and ran for $10^{5}$ iterations. For the DIS-SAM$^{*}$ model trained on the HQSeg-44k dataset, the number of training iterations was increased to $2*10^{5}$. Other training configurations remained the same as those of the DIS-SAM model. The experiments were done on an RTX 4090 GPU.

\begin{figure*}
  \centering
  \includegraphics[width=0.96\textwidth]{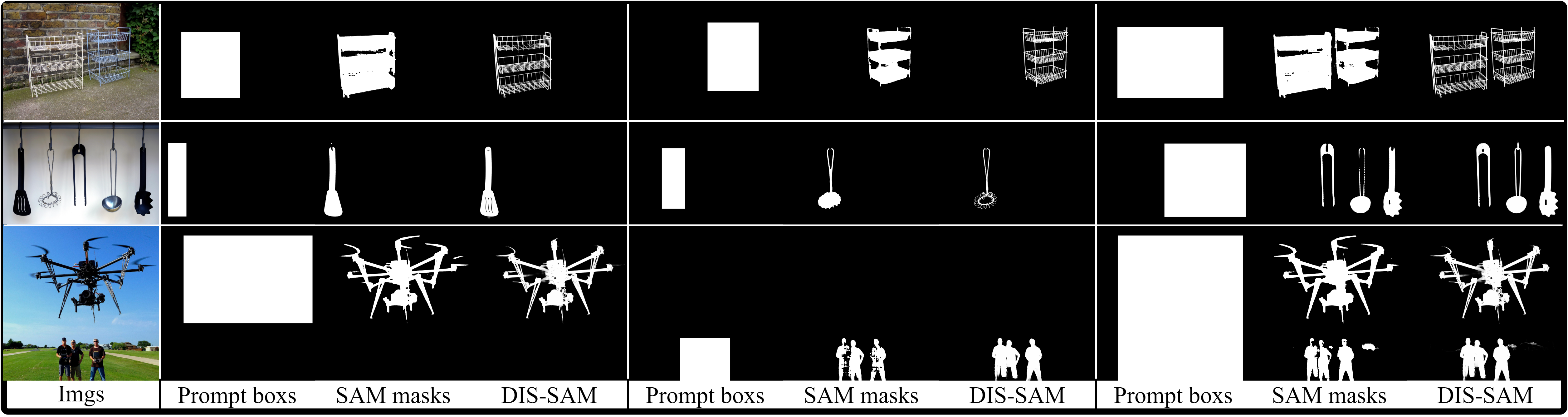}
  \caption{Visual results of promptable capability of DIS-SAM.}
  \vspace{-0.35cm}
  \label{fig:prompt}
\end{figure*}

\subsection{Comparison with State-of-the-Arts}
We compare DIS-SAM with SAM (2023) \cite{kirillov2023segment}, HQ-SAM (2023) \cite{ke2024segment}, Pi-SAM (2024) \cite{mengzhen2024segment}, original IS-Net (2022) \cite{qin2022highly}, UDUN (2023)\cite{pei2023unite} and BiRefNet (2024)\cite{zheng2024birefnet}. 
Note that IS-Net, UDUN and BiRefNet do not have prompt box or mask input, whereas SAM, HQ-SAM, Pi-SAM, and DIS-SAM take the original image and prompt box as inputs. 
The model weight parameters of SAM and HQ-SAM were their original ones without fine-tuning on the DIS task. The backbones of SAM, HQ-SAM, Pi-SAM, and DIS-SAM all adopt ViT-B\cite{dosovitskiy2020vit}.

As shown in \tabref{tab:table1}, one can clearly observe that, despite of its simplicity, DIS-SAM significantly outperforms the rest models across all test sets.
Compared to original IS-Net, DIS-SAM achieves a notable improvement, especially in $F^{max}_\beta$, thanks to the incorporation of prompt boxes and SAM's coarse masks. For associated visual comparisons, we provide them in \figref{fig:compare}. DIS-SAM is capable of segmenting more details. 
Compared to DIS-SAM, as DIS-SAM$^{*}$ is trained on a larger dataset but with lower quality, DIS-SAM$^{*}$ generally performs worse on the DIS-5K dataset than DIS-SAM in \tabref{tab:table1}. For more results, please refer to Supplemental Material.

\begin{table}
\centering
\caption{Following HQ-SAM, zero-shot generalization performance comparisons of DIS-SAM with SAM and HQ-SAM. The symbols ↑/↓ indicate that higher/lower scores are better.}
\label{tab:table3}
\small
\renewcommand{\arraystretch}{1}
\renewcommand{\tabcolsep}{0.5mm}
\begin{tabular}{cr|cccc}
            \toprule
            Test Dataset & Metric & SAM & HQ-SAM & \textbf{DIS-SAM} & \textbf{DIS-SAM$^{*}$} \\
            \hline
            \multirow{5}{*}{COIFT \cite{liew2021deep}}
            & $F^{max}_\beta\uparrow$ & 0.966 & 0.974 & 0.982 & \textbf{0.986} \\
            ~ & $F^w_\beta\uparrow$ & 0.967 & \textbf{0.976} & 0.969 & 0.969 \\
            ~ & $~M~\downarrow$ & 0.007 & \textbf{0.005} & \textbf{0.005} & 0.006 \\
            \multirow{2}{*}{(280 samples)}  ~ & $S_{\alpha}\uparrow$ & 0.964 & 0.971 & 0.978 & \textbf{0.982} \\
            ~ & $E_{\phi}^{m}\uparrow$ & 0.988 & \textbf{0.991} & 0.988 & 0.987 \\
            ~ & $HCE_\gamma\downarrow$ & 31 & 30 & \textbf{14} & 16 \\
            \hline
            \multirow{5}{*}{HRSOD \cite{zeng2019towards}} 
            & $F^{max}_\beta\uparrow$ & 0.952 & 0.965 & 0.971 & \textbf{0.974} \\
            ~ & $F^w_\beta\uparrow$ & 0.939 & \textbf{0.956} & 0.953 & 0.949 \\
            ~ & $~M~\downarrow$ & 0.013 & 0.009 & \textbf{0.008} & \textbf{0.008} \\
            \multirow{2}{*}{(287 samples)}   & $S_{\alpha}\uparrow$ & 0.947 & 0.958 & \textbf{0.969} & 0.904 \\
            ~ & $E_{\phi}^{m}\uparrow$ & 0.976 & \textbf{0.984} & \textbf{0.984} & 0.982 \\
            ~ & $HCE_\gamma\downarrow$ & 317 & 294 & \textbf{188} & 216 \\
            \hline
            \multirow{5}{*}{ThinObject5K \cite{liew2021deep}} 
            & $F^{max}_\beta\uparrow$ & 0.859 & 0.934 & 0.933 & \textbf{0.968} \\
            ~ & $F^w_\beta\uparrow$ & 0.836 & 0.919 & 0.899 & \textbf{0.943} \\
            ~ & $~M~\downarrow$ & 0.089 & 0.035 & 0.039 & \textbf{0.021} \\
            \multirow{2}{*}{(500 samples)}   & $S_{\alpha}\uparrow$ & 0.836 & 0.907 & 0.908& \textbf{0.938} \\
            ~ & $E_{\phi}^{m}\uparrow$ & 0.879 & 0.947 & 0.939 & \textbf{0.966} \\
            ~ & $HCE_\gamma\downarrow$ & 395 & 321 & 218 & \textbf{211} \\
            \hline
            \multirow{5}{*}{ALL} 
            & $F^{max}_\beta\uparrow$ & 0.899 & 0.929 & 0.953 & \textbf{0.963} \\
            ~ & $F^w_\beta\uparrow$ & 0.881 & 0.920 & 0.925 & \textbf{0.929} \\
            ~ & $~M~\downarrow$ & 0.044 & 0.023 & 0.021 & \textbf{0.018} \\
            \multirow{2}{*}{(1,067 samples)}   & $S_{\alpha}\uparrow$ & 0.889 & 0.921 & \textbf{0.941} & 0.933 \\
            ~ & $E_{\phi}^{m}\uparrow$ & 0.933 & 0.960 & 0.965 & \textbf{966} \\
            ~ & $HCE_\gamma\downarrow$ & 565 & 508 & \textbf{352} & 372 \\
            \bottomrule
\end{tabular}
\vspace{-0.35cm}
\end{table}

\subsection{Zero-shot Evaluation}
According to the zero-shot evaluation in \tabref{tab:table3}, although DIS-SAM was trained only on the DIS-5K dataset, it practically surpasses HQ-SAM on most datasets.
Notably, as ThinObject5K is an artificial dataset, where an object is directly placed in the center of an image, its composition is quite different from those natural images. Since the training set of DIS-SAM consists of only natural images, the results of DIS-SAM on this dataset are less promising and generally worse than HQ-SAM. However, it still outperforms SAM.
Furthermore, one can see that DIS-SAM$^{*}$, which was trained on the same data set as HQ-SAM, outperforms HQ-SAM on all data sets. The above remarkable results demonstrate that the proposed DIS-SAM framework shows good generalizability.
We present examples in \figref{fig:prompt} demonstrating that DIS-SAM inherits the promptable capability of SAM. Different from other DIS methods that can only segment primary objects without interactivity, DIS-SAM can segment fine-grained results which dynamically update in response as the prompt box is adjusted. This highlights the advantages of leveraging the prompt box, SAM masks, and the ground truth enrichment. For more results, please refer to Supplemental Material.

\begin{table}
    \centering
    \caption{Ablation results on dataset DIS-VD. Notation ``Enrich.'' means whether data enrichment is deployed. ``Box'' and ``Mask'' indicate whether to concatenate prompt box or SAM's coarse mask as input during the second stage.}
    \label{tab:table2}
    \small
    \renewcommand{\arraystretch}{1}
    \renewcommand{\tabcolsep}{0.5mm}
        \begin{tabular}{ccc|cccccc}
            \toprule
            Enrich& Box & Mask& $F^{max}_\beta\uparrow$ &$F^w_\beta\uparrow$ &$~M~\downarrow$ &$S_{\alpha}\uparrow$ &$E_{\phi}^{m}\uparrow$ &$HCE_\gamma\downarrow$\\
            \hline
            --& --& --& 0.791& 0.717& 0.074& 0.813& 0.856& 1116\\
            --& \checkmark& --& 0.901& 0.829& 0.042& 0.891& 0.821& 1039\\
            --& --& \checkmark& 0.910& 0.850& 0.037& 0.822& 0.935& 1028\\
            --& \checkmark& \checkmark& 0.913& 0.874& 0.032& \textbf{0.913}& \textbf{0.948}& 1010\\
            \checkmark& \checkmark& \checkmark& \textbf{0.920}& \textbf{0.877}& \textbf{0.031}& 0.909& \textbf{0.948}& \textbf{987}\\
            \bottomrule
        \end{tabular}
\end{table}

\subsection{Ablation Study}
\begin{figure*}
  \centering
  \includegraphics[width=0.96\textwidth]{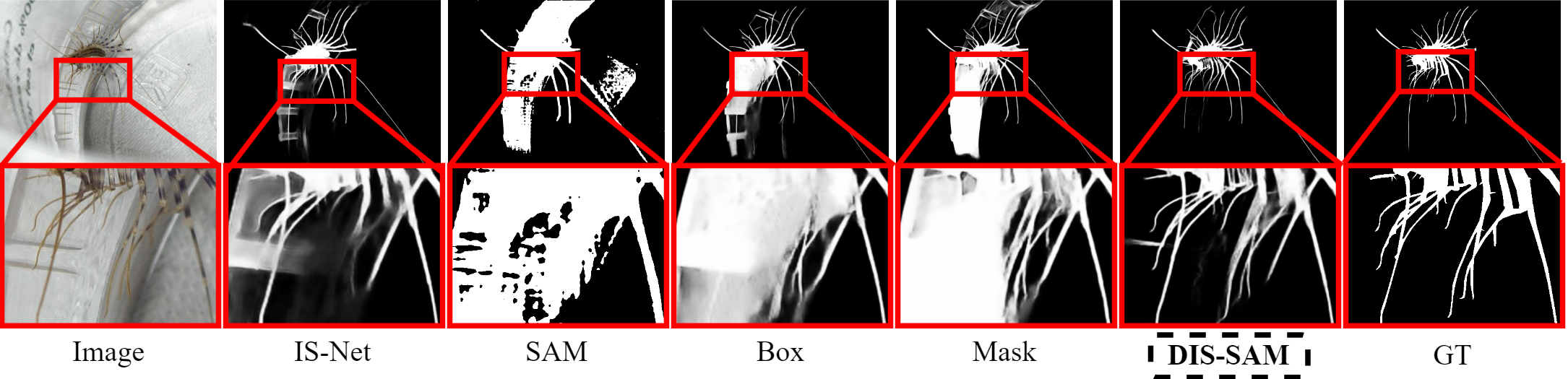}
  \caption{Visual results of ablation study. ``Box'' and ``Mask'' indicate whether to concatenate prompt box or SAM's mask as input during the second stage.}
    \vspace{-0.35cm}
  \label{fig:abaltion}
\end{figure*}

\tabref{tab:table2} shows the ablation results on the DIS-VD dataset with 470 samples, evaluating the effects of data enrichment, the prompt box, and the SAM mask. 
For the settings without prompt box and SAM mask, except that the first row of \tabref{tab:table2} indicates the results of the original IS-Net, we use all-zeros black images as input, to keep the five-channel input form consistent.  
The results show that the prompt box greatly improves object localization, improving metrics such as $F^{max}_\beta$. Meanwhile, the SAM mask reduces $HCE_\gamma$, offering complementary information for fine-grained segmentation. Combining them boosts performance across all metrics. Also, data enrichment further improves overall accuracy, showing its role in enhancing segmentation precision and robustness.
\figref{fig:abaltion} shows the visualizations of some ablation experiments, including using only the prompt box or SAM mask. When the original IS-Net is adopted, the network fails to capture the target accurately. When using only the box, the model lacks the boundary guidance from the mask, resulting in a failure to recognize the target. When using only the mask, the model may make errors in target segmentation due to mask errors. For more results, please refer to Supplemental Material.

%

\section{Conclusion}

We propose DIS-SAM, a novel two-stage framework that integrates the SAM with a modified IS-Net, specifically designed to achieve highly detailed DIS. The framework builds upon SAM's promptable nature and combines it with the powerful capabilities of IS-Net to refine object boundaries and enhance segmentation accuracy. Experimental results show that DIS-SAM significantly outperforms the original IS-Net on the DIS-5K dataset, achieving higher segmentation accuracy and precision, particularly in delineating fine details and object contours. While the two-stage approach of DIS-SAM demonstrates considerable improvements in segmentation quality, it introduces some redundancy due to the involvement of both SAM and IS-Net. Future research could focus on developing more streamlined, one-stage architectures that eliminate redundancy while maintaining or even enhancing performance.

\bibliographystyle{IEEEbib}
\bibliography{icme2025references}

\end{document}